\documentclass[10pt,twocolumn,letterpaper]{article}

\usepackage[pagenumbers]{cvpr} 

\usepackage{times}   
\usepackage{epsfig}
\usepackage{amsmath}
\usepackage{amssymb}
\usepackage{booktabs}
\usepackage{algorithm}
\usepackage{microtype}
\usepackage{mathtools}
\usepackage{comment}
\usepackage{dsfont}
\usepackage{amsthm}
\usepackage{cite}
\usepackage{xcolor,colortbl}
\usepackage{enumitem}

\def\Vec#1{{\boldsymbol{#1}}}

\definecolor{blond}{rgb}{0.98, 0.94, 0.75}

\definecolor{cvprblue}{rgb}{0.21,0.49,0.74}
\usepackage[pagebackref,breaklinks,colorlinks,citecolor=cvprblue]{hyperref}

\def\paperID{1001} 
\def\confName{CVPR}
\def\confYear{2024}

\title{Hyper-VolTran: Fast and Generalizable \\ One-Shot Image to 3D Object Structure via HyperNetworks}

\author{%
\vspace{0.3cm}
  Christian Simon\and Sen He\and Juan-Manuel P\'erez-R\'ua\and Mengmeng Xu\\
  \hspace{-12cm} Amine Benhalloum\quad\quad Tao Xiang \\
  {\hspace{-11.5cm} Meta}\\
}

\begin{document}
\maketitle

\begin{abstract}
Solving image-to-3D from a single view 
is an ill-posed problem, and current neural reconstruction methods addressing it through diffusion models still rely on scene-specific optimization, constraining their generalization capability.
To overcome the limitations of existing approaches regarding generalization and consistency, we introduce a novel neural rendering technique. Our approach employs the signed distance function (SDF) as the surface representation and incorporates generalizable priors through geometry-encoding volumes and HyperNetworks. Specifically, our method builds neural encoding volumes from generated multi-view inputs. 
We adjust the weights of the SDF network conditioned on an input image at test-time to allow model adaptation to novel scenes in a feed-forward manner via HyperNetworks.
To mitigate artifacts derived from the synthesized views, we propose the use of a volume transformer module to improve the aggregation of image features instead of processing each viewpoint separately. Through our proposed method, dubbed as \textbf{Hyper-VolTran}, we avoid the bottleneck of scene-specific optimization and maintain consistency across the images generated from multiple viewpoints. Our experiments show the advantages of our proposed approach with consistent results and rapid generation.

\end{abstract}

\section{Introduction}
\label{sec:intro}

\begin{figure}[t]
    \centering
    \includegraphics[width=0.49\textwidth]{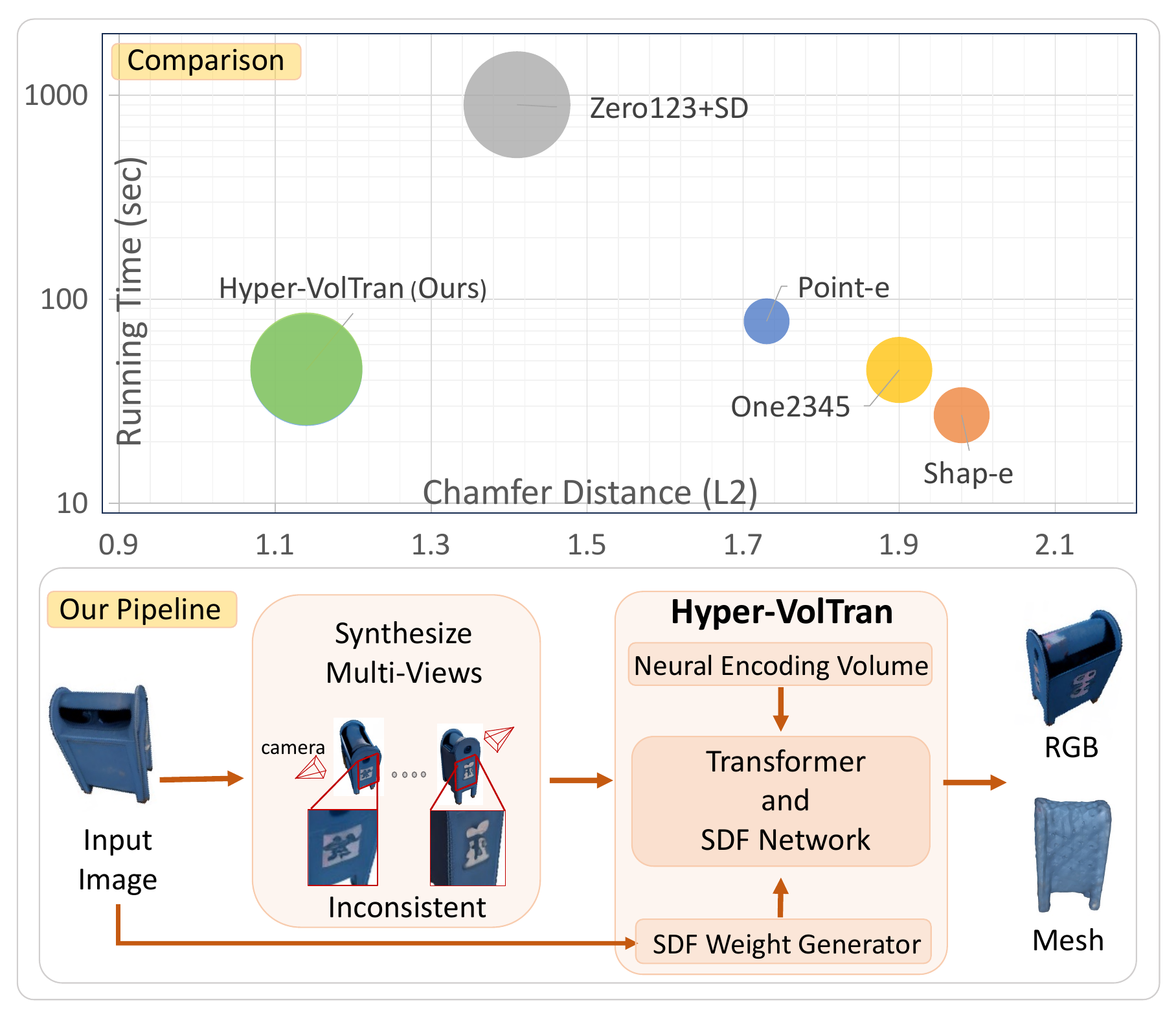}
    \caption{ \textbf{Top}: Comparison of our proposed method against baselines on the running time and Chamfer Distance with the bubble area indicating IoU. \textbf{Bottom}: 
    Our pipeline comprises two components for image-to-3D by synthesizing multi-views from a diffusion model and mapping from multi-views to SDFs using an SDF network with weights generated from a HyperNetwork. 
    }
    \label{fig:intro}
\end{figure}
\vspace{-0.3cm}


Recent progress in neural 3D reconstruction has brought significant implications in various applications, \eg, novel view synthesis~\cite{williams20213drecon,barron2021mipnerf,xie2023snerf,t2023is}, and robotic vision~\cite{simeonov2022relational,wen2023bundlesdf,nerfrpn,lewis2022narf}. Specifically, there has been a growing interest in neural fields~\cite{xie2023snerf,yu2020pixelnerf,barron2021mipnerf} to extract 3D information from multiple images given known camera parameters. NeRF~\cite{mildenhall2020nerf} and the Signed Distance Function (SDF)~\cite{lior2021sdf} are image-to-3D reconstruction techniques~\cite{barron2021mipnerf,chen2021mvsnerf,wang2021neus} that produce plausible geometry and, hence, novel views. Despite great progress, 
achieving accurate 3D object reconstruction via
neural implicit methods~\cite{barron2021mipnerf,mildenhall2020nerf,lior2021sdf} still requires a substantial number of images featuring consistent view and appearance and precise camera poses to reconstruct 3D objects accurately.

In fact, collecting data from multiple views might not always be feasible when the resources are limited. Several works~\cite{long2022sparseneus,yu2020pixelnerf,chen2021mvsnerf} demonstrate a capability to mitigate issues on 3D reconstruction under a sparse set of images. One key technique in these approaches is to build neural encoding volume projected from multiple input views. Though these techniques can perform on limited inputs, reconstructing 3D from a single image remains challenging and requires a strong prior to enabling the neural reconstruction model to produce plausible shapes and colors of unseen perspectives.


A recent development in generative models~\cite{saharia2022imagen,liu2023zero123,yu2023cm3leon,cong2023flatten} has shown promising results in 2D image generation that can act as a strong prior for unseen perspectives. Several works approach this problem using the guidance of a diffusion model~\cite{poole2022dreamfusion}. In particular, Poole \etal~\cite{poole2022dreamfusion} introduce Score Distillation Sampling (SDS)~\cite{poole2022dreamfusion} in which the neural reconstruction model learns through the feedback error from a diffusion model. The diffusion model is frozen without any updates while the NeRF~\cite{mildenhall2020nerf} weights are updated during optimization. Even though this technique is capable of reconstructing 3D scenes, per-scene optimization is still required, which usually takes up to 1 hour to converge on a single GPU. This constraint restricts the practicality of this approach, particularly when it comes to efficiently performing 3D reconstruction. To achieve fast 3D reconstruction, a generalized prior that allows one feed-forward operation through the networks is required instead of relying on an expensive per-scene optimization.


An alternative method for rapid 3D reconstruction is to utilize a diffusion model and synthesize multi-view images. This can be achieved by leveraging a diffusion model that can produce images based on slight variations in camera parameters~\cite{liu2023zero123}. Nevertheless, creating images using a multi-view image generator (\eg, Zero123~\cite{liu2023zero123}) can be challenging in terms of preserving geometry consistency. Rather than optimizing a network for each object as in~\cite{poole2022dreamfusion}, we aim to preserve only one network to generalize for many objects. To achieve this, we can exploit neural encoding volume built from the projection of image features with known camera parameters as in~\cite{chen2021mvsnerf,wang2021neus,long2022sparseneus}. While these approaches show promise, they still suffer from suboptimal results when employed for 3D reconstruction involving unseen objects. 

 In this work, we aim to address the aforementioned challenges, focusing on generalization, speed, and inconsistency issues. To this end, we introduce a neural network to address these concerns by employing an SDF network generated by HyperNetworks~\cite{ha16hypernet} and a Volume Transformer (VolTran) to alleviate the impact of inconsistent examples. Our approach explores the potential for generalization by introducing a latent variable obtained from an image encoder (\eg, CLIP~\cite{radford2021clip}) to yield image representations. Subsequently, we employ these image representations to generate the weights of the SDF, addressing the challenge of generalization. Please see Fig.~\ref{fig:intro} (bottom) for an illustration of our technique.
 To summarize, our contributions include:
\begin{enumerate}
    \item We propose a generalizable prior for 3D mesh reconstruction with a few synthesized data by assigning the weights of SDFs based on the input image embedding. 
    \item We propose a transformer module for aggregation to enable working on inconsistent shapes and colors across different viewpoints.  
    \item We also show that our method only requires one feed-forward process and comfortably constructs a 3D mesh with negligible additional processing time $\sim$5 seconds.
\end{enumerate}

\section{Related Work}
\label{sec:related_work}


\paragraph{Diffusion models for 2D to 3D reconstruction.} Reconstructing a full 3D structure from only a few 2D images is challenging due to the inherent ill-posedness of the problem. 
However, recent advances in generative models and, in particular, diffusion models provide a promising direction toward obtaining the priors about the 3D world 
that are necessary to reconstruct the full 3D structure of an object from a single image. 
For example, they are used as an indirect way to provide feedback during the image-to-3D reconstruction process in~\cite{poole2022dreamfusion,tang2023makeit3d,haochen2022sjc,chen2023fantasia3d,melaskyriazi2023realfusion}. 
A notable work so-called DreamFusion~\cite{poole2022dreamfusion} 
proposes text-to-3D generation by Score Distillation Sampling (SDS), which allows optimization-guided generation of NeRF-parametrized~\cite{mildenhall2020nerf} 3D scenes.
A concurrent work using Score Jacobian Chaining~\cite{haochen2022sjc} uses a similar approach, exploiting the chain rule on the outputs of a pretrained image generation model. 
Tang \etal~\cite{tang2023makeit3d} extend the idea with coarse and refining stages to enhance the outputs with textured point clouds.  Recently, Zero123~\cite{liu2023zero123} describes a diffusion model that takes an input image and camera parameters to synthesize a novel view. This model can generate more consistent multi-view images compared to an off-the-shelf diffusion model like Imagen~\cite{saharia2022imagen}. 
Albeit a 
promising direction to reconstruct 3D models, per-scene optimization is still required and the neural implicit function is limited to represent only one object. Thus, the generalization of the trained model is  limited for unseen objects. 

\vspace{-0.46cm}
\paragraph{Generalizable priors for fast 3D reconstruction.}  An ideal implementation of 3D reconstruction is a single model that can generalize to unseen objects, enabling 3D generation using a forward-pass approach only without applying further per-scene optimization.
PixelNeRF~\cite{yu2020pixelnerf} as a pioneer work in this direction proposes to extract feature volumes from an input image which are then passed through a NeRF model along with the camera extrinsic parameters. Chen \etal~\cite{chen2021mvsnerf} present an approach called MVSNeRF using cost volumes built of warped 2D image features and then regress volume density with a pass through an MLP (\ie, neural encoding volumes) as the base geometry. Then, the neural encoding volume is used as an additional input to the NeRF model. SparseNeus~\cite{long2022sparseneus} extends MVSNeRF~\cite{chen2021mvsnerf} to work on a few-data regime by proposing cascaded geometry reasoning to refine the details of a 3D object. 
However, this approach still requires multi-view inputs, 
with no obvious mechanism to extend it to a single image.
To tackle the problem of 3D reconstruction from a single image, Liu \etal~\cite{liu2023one2345} propose a method called One2345 to exploit a diffusion model (\eg, Zero123~\cite{liu2023zero123}) to generate some example images with estimated camera poses. To improve the precision of the reconstructed geometric models, One2345~\cite{liu2023one2345} employs SDFs~\cite{lior2021sdf} rather than NeRFs~\cite{mildenhall2020nerf}. 
The challenge of this approach is inconsistency in generated examples, making it difficult to reconstruct 3D scenes that fully respect the input appearance. 

Another approach for avoiding per-scene optimization is to train a large-scale model with self-supervised learning and make use of large-scale labeled text-to-3D data. Point-e~\cite{nichol2022pointe}, a system to generate 3D point clouds from text description, is a pioneer in this direction. Following up this work, Shap-e~\cite{jun2023shape}  directly generates the weights of the neural implicit model that can be rendered as meshes and radiance fields. This method generates multiple synthetic images then a neural 3D reconstruction technique (\eg, SDF~\cite{lior2021sdf} or NeRF~\cite{mildenhall2020nerf}) is employed to produce 3D models. This model cuts the cost of image-to-3D reconstruction from several GPU hours to 1-2 minutes. While this method can produce results quickly, the quality of the reconstructed 3D surfaces remains subpar.
Unlike all these prior works, our proposed method can generate accurate 3D reconstruction with competitive processing time (\ie, less than 1 minute).

\vspace{-0.32cm}
\paragraph{Context-based learning.} 
In few-shot learning, the concept of leveraging contextual information for achieving optimal performance across diverse input conditions is a well-established idea, as indicated by previous works like ~\cite{simon2022multilabel,ha16hypernet,finn2017maml,simon2020modulate,simon2020subspace,xu2022where}. Some of these methods involve model parameter updates through gradient descents, exemplified by several works~\cite{zintgraf2018cavia,finn2017maml}. However, these approaches still require multiple feed-forward operations to update the model.
Our focus lies in developing an approach that accomplishes context understanding with just a single feed-forward operation, without the need for additional optimization steps. To achieve this, we opt to adopt context-based information by generating neural network weights. Specifically, we draw inspiration from HyperNetworks~\cite{ha16hypernet} designated to generate neural network weights based on the provided context. 

\vspace{-0.1cm}
\section{Proposed Method}
\label{sec:proposed_method}

\begin{figure*}[t]
    \centering
    \includegraphics[width=1.0\textwidth]{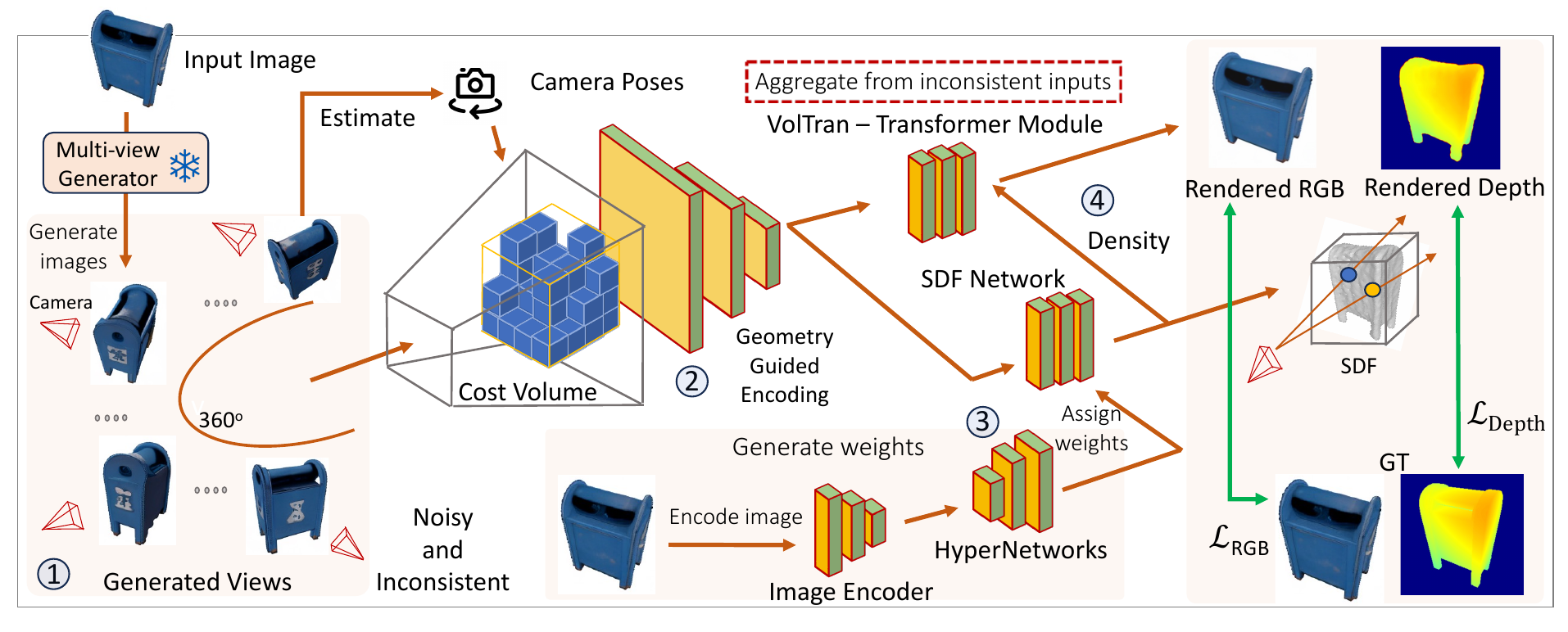}
    \caption{\textbf{Our training pipeline starts from a single image. Expanding a single view to an image set using a viewpoint-aware generation model, our method employs supervised learning with RGB and depth regression losses.}
    Specifically, 
    1) Utilizing $N$ RGB images and depth maps, we generate additional viewpoints and camera poses.
    2) Geometry-Guided Encoding is derived from warped image features in the form of a Cost Volume.
    3) Instead of test-time optimization, we obtain SDF weights with a single pass of a HyperNetwork module, considering image appearance through visual encoding.
    4) The geometry-encoded volume and the image features are passed to the SDF network and a transformer module to reveal the complete 3D object structure. 
    Hence, our method Hyper-VolTran encompasses quick adaption to novel inputs thanks to our HyperNetwork design and consistent structures from global attention. 
}
\label{fig:pipeline}
\end{figure*}

Our 3D neural reconstruction pipeline has two streams, as shown in Fig.~\ref{fig:pipeline}.
Given a single-view image and its depth map, we first synthesize multi-view images via a diffusion model.
Then, as shown in the upper stream of the figure, the synthesized images are fed into a neural encoding volume to obtain the 3D geometry representation of its structure.
The geometry representation is combined with the images to predict a rendered RGB map by our proposed transformer module, VolTran.
Meanwhile, we also use the synthesized multi-view images in a HyperNetwork to estimate an SDF weight, shown in the bottom stream. The SDF network predicts SDFs for surface representations that will later be used for rendering the depth map and extracting the mesh.
Therefore, we name our approach \textbf{Hyper-VolTran}. 


\subsection{{One to multiple-view images}} \label{sec:multiview}
We begin our pipeline by leveraging a pretrained generative model. This enables us to expand a single input image into multiple views from a broader set of object viewpoints, albeit with some imperfections.
For fair comparison, we strictly follow the approach outlined in~\cite{liu2023zero123} to leverage elevation and azimuth conditioning.
\paragraph{Synthesized views.} 
Given a single RGB image and its corresponding depth map denoted as $\Vec{I} \in \mathbb{R}^{H\times W \times 3}$,  and $\Vec{D} \in \mathbb{R}^{H\times W}$, respectively, {we 
follow} Zero123~\cite{liu2023zero123} to normalize its shape and use a spherical camera system {for the depth map}. 
{We apply an off-the-shelf image generation model to create}
$N$ RGB images and depth maps sampled uniformly from several viewpoints {according to} ground-truth camera parameters~\cite{liu2023one2345}. 
{Concretely} for training, we form a set of RGB images and depth maps of an object as the source set $\mathcal{I} =\{\Vec{I}_1, \cdots , \Vec{I}_N \}$ and $\mathcal{D} =\{\Vec{D}_1, \cdots , \Vec{D}_N\}$. {Note that} both RGB and depth images are used as {training targets} to supervise the model in the training stage. However, those depth maps are omitted in the testing phase. 

\subsection{Geometry-Aware Encoding}
Geometry-aware encoding is essential in building a generalized method for surface prediction from multi-view images. {Our approach employs}
neural encoding volumes~\cite{yao2018mvsnet,chen2021mvsnerf} to construct 3D geometry based on the {diversified} input views from Sec.~\ref{sec:multiview} and their associated camera poses. To this end, we warp 2D image features from the $N$ input images onto a localized plane situated within the reference view's frustum.

\paragraph{Neural encoding volume.}
In deep multi-view stereo~\cite{yao2020blendedmvs,yao2018mvsnet}, 3D geometry can be inferred in the form of Cost Volume construction. Let $f_\theta:\mathbb{R}^{H\times W \times 3} \rightarrow \mathbb{R}^{ H\times W \times C}$ be the mapping from an input image to a feature map. Similar to~\cite{long2022sparseneus,yao2018mvsnet}, {we encode} images using
a Feature Pyramid Network~\cite{he2017pyramid} {as the mapping function} to extract a neural feature map, \ie, $\Vec{F}_i =  f_\theta(\Vec{I}_i)$. {Besides,} we partition the scene's bounding volume into a grid of voxels. Then, along with the intrinsic and extrinsic camera parameters $\Vec{P}=[\Vec{K}, \Vec{R}, \Vec{t}]$ for each image $\Vec{I}_i$, the neural feature map is projected based on 
each vertex $v$, {and the output is} denoted as $\Vec{F}_{i}({\Pi_i(\Vec{v})})$, where ${\Pi_i(\Vec{v})}$ projects $\Vec{v} \in \mathbb{R}^3$ onto the local plane by applying $\Vec{P}$~\cite{yao2018mvsnet}. In particular, the homography warping is applied for each view $i$, and the final neural encoding volume $\Vec{G}$ can be computed as Eq.~\ref{eq:CostVolume}.
\begin{equation}
    \Vec{G} = 
    \phi\Big(\textrm{Var}\big(\{\Vec{F}_{i}({\Pi_i(\Vec{v})})\}_{i=1}^{N}\big)\Big).
    \label{eq:CostVolume}
\end{equation}
Here $\textrm{Var}(\{\Vec{F}_{i}({\Pi_i(v)})\}_{i=0}^{N-1})$ is the Cost Volume, $\textrm{Var}$ {means} the variance {over} $N$ viewpoints, and $\phi$ denotes a function responsible for regularizing and propagating scene information {instantiated as} a sparse 3D CNN {(\ie, Geometry Guided Encoding)}. {Since} the variance accommodates differences in the image appearance among multiple input perspectives, $\Vec{G}$ acquires the ability to encode {complex} 3D scene geometry and appearance from diversified images. {Thus,} these volume features contain appearance-aware information that can be later used for volume rendering and SDF predictions.

\subsection{Volume Rendering}
{A neural encoding volume previously computed is employed to predict both the density and view-dependent radiance at arbitrary locations within a scene.} Next, this facilitates the utilization of differentiable volume rendering to predict the colors of images. 
For volume rendering, we opt to use SDF~\cite{lior2021sdf} instead of NeRF~\cite{mildenhall2020nerf} for a more accurate surface reconstruction.

\paragraph{Signed Distance Function (SDF).}
SDFs represent 3D surfaces using a positional function that provides the nearest distance to the surface.
{Given} an arbitrary 3D location {in our} setup, {we use} an MLP $f_{\Psi}:\mathbb{R}^{d} \rightarrow \mathbb{R}$ {as an SDF} to represent 3D surfaces. {Although the generic SDF input has $d=3$ as} the signed distance is associated with a point $\Vec{z} \in \mathbb{R}^3$, our method uses a higher $d$ as the input consists of the concatenation of feature from neural encoding volumes, colors, and image features. 
{Another limitation of the generic SDF is the lack of generalization ability. For example,}
when using the neural encoding volume as an input, we can train an SDF network on {a large collection of} 3D objects~\cite{chen2021mvsnerf,long2022sparseneus} {to} avoid per-scene optimization. {In testing, however, the SDF network is usually frozen}~\cite{liu2023one2345,long2022sparseneus} and limited to the known objects. We propose a more {adaptable} approach to dynamically assign MLP's weights based on the generated outputs of a HyperNetworks~\cite{ha16hypernet}, which is conditioned on the input image. 

\paragraph{HyperNetworks for an SDF network.}
HyperNetworks~\cite{ha16hypernet} constitute a neural model that generates the weights for a target network designed to generalize on various tasks given a context. Rather than preserving a neural network fixed during test time, HyperNetwork offers a mechanism to assign weights based on a condition dynamically. 
{Mathematically, we} design a HyperNetwork module $\delta_l(.)$ {to produce the weight} for each layer ${\psi}_l$  of the SDF network $f_\Psi$:
\begin{equation}
     {\psi}_l = \delta_l(\xi(\Vec{I}_1)).
\end{equation}
To encode the input image, we use a {pretrained image encoder} $\xi$ that {reduces the image dimensionality} from RGB space to a latent space. 
Unlike the past work~\cite{erkocc2023hyperdiffusion} that needs to optimize neural networks for every single object, our method trains the module on the fly without requiring per-scene optimization and directly calculating losses between two neural network parameters. 
{Since our condition is the feature representation of the input object, our HyperNetwork can produce a more dedicated and appropriate weight for its target network.}
{On the other hand, as} we utilize {the} output of the Hypernetwork~\cite{ha16hypernet} to assign weights to the {SDF network}, {our model generalizes better on the new object during inferences, especially when the object shares similar semantics with the training data}. 
Moreover, the hypernetworks are directly updated with a loss from RGB and depth map in our pipeline. Thus, {we do not have to store the individual} optimal weight parameter after per-scene optimization.

\paragraph{Rendering from SDFs.}
To estimate the parameters of the neural SDF and color field, we adopt a volume rendering method from NeuS~\cite{wang2021neus} to render colors and volumes based on the SDF representations. For a given pixel, we describe $M$ emitted rays from that pixel as $\{\Vec{p}(t) = \Vec{o} + t\Vec{v} | t \geq 0\}$, with $\Vec{o}$ being the camera's focal point and $r$ representing the ray's unit direction. 
 We feed the combined features through an MLP and employ the \textit{softmax} function to derive the blending weights denoted as $\{\omega_i\}^N_{i=1}$. The radiance at a given point $\Vec{p}$ and viewing direction $\Vec{v}$ is calculated as the weighted sum in Eq~\ref{eq:hat_c}.
\begin{equation}
    \Vec{\hat{c}} = \sum^{N}_{i=1}\omega_i.\Vec{c}_i,
    \label{eq:hat_c}
\end{equation}
where $\Vec{c}_i$ is the color of source view $i$.  
{Given the radiance, our volume rendering strategies is expressed in Eq~\ref{eq:color_render}.}
\begin{align}
\label{eq:color_render}
\Vec{\hat{C}} &= \sum^{M}_{j=1} T_j \alpha_j \Vec{\hat{c}}_j, \\
\alpha_j  &= 1 - \exp[{- \int_{t_j}^{t_{j+1}} \rho(t)dt}].
\end{align}
Here, $T_j = \prod^{j=1}_{k=1}(1-\alpha_k)$ is a discrete accumulated transmittance, $\alpha_k$ is the discrete opacity, and $\rho(t)$ denotes opaque density.
{The} rendered depth {map} can be derived as Eq.~\ref{eq:depth}:
\begin{equation}
\Vec{\hat{D}} = \sum^{M}_{j=1} T_j \alpha_j t_j.
\label{eq:depth}
\end{equation}
{Note the rendering process is fully differentiable; we train the pipeline in a supervised manner so that the model can predict the rendered colors $\Vec{\hat{C}}$ and depths $\Vec{\hat{D}}$ in inference.}




\paragraph{\textit{VolTran}: {multi-view aggregation} transformer.}

Pixel data is inherently confined to a local context and lacks broader contextual information, frequently leading to inconsistent surface patches, particularly in the case of sparse input data. {One trivial solution is} to aggregate features across different views to capture {the} projected features from {multiple} views. {Unfortunately,} the synthesized {views} might be corrupted due to {the flaws in the generative} model, a simple aggregation~\cite{long2022sparseneus,liu2023one2345,yao2018mvsnet} (\eg, average and max. pooling) might fail to render shapes and colors accurately.  
We propose a transformer module called \textit{VolTran} based on the self-attention design in~\cite{vaswani2017attention} to encode global information from different $N$ viewpoints. 
{Besides} the inputs, we learn an aggregation token as {an extra token to obtain} a corresponding output for a target view. Formally, let $\Vec{X} \in \mathbb{R}^{N+1\times d}$ be a matrix with rows composed of the tokens from source views and the aggregation token by concatenating the feature from color $\Vec{c}_i$, image feature $\Vec{F}_i(\Pi(\Vec{v}))$, and volume feature $\Vec{G}$ yielding the dimension $d$. We denote $f_V(.), f_Q(.), f_K(.)$ as functions to map values, queries, and keys of a transformer module. {Thus, the} aggregation operation can be calculated by the self-attention module, as shown in Eq.~\ref{eq:sa}:

\begin{equation}
    \textrm{Attn}(\Vec{X}) = \textrm{Softmax}(\Vec{A})f_V(\Vec{X}),
    \label{eq:sa}
\end{equation}
where $\Vec{A}_{i,j} = f_Q(X_i)^\top f_K(X_j) / \gamma$ for all $ i, j \in [N] $.
As we apply multi-head attention, it can be formulated as $\textrm{MHA}(\Vec{X}) = [\textrm{Attn}_1(\Vec{X}), \cdots, \textrm{Attn}_3(\Vec{X})]\Vec{W}_H$. We opt to use LayerNorm to normalize the intermediate activations and skip connection to stabilize training. The final 
output from the transformer module, an MLP, is introduced as a mapping function to obtain the blending weight $\omega_i$. Afterwards, the final color can be obtained as in the SDF rendering pipeline. 

\begin{figure*}[t]
    \centering
    \includegraphics[width=0.99\textwidth]{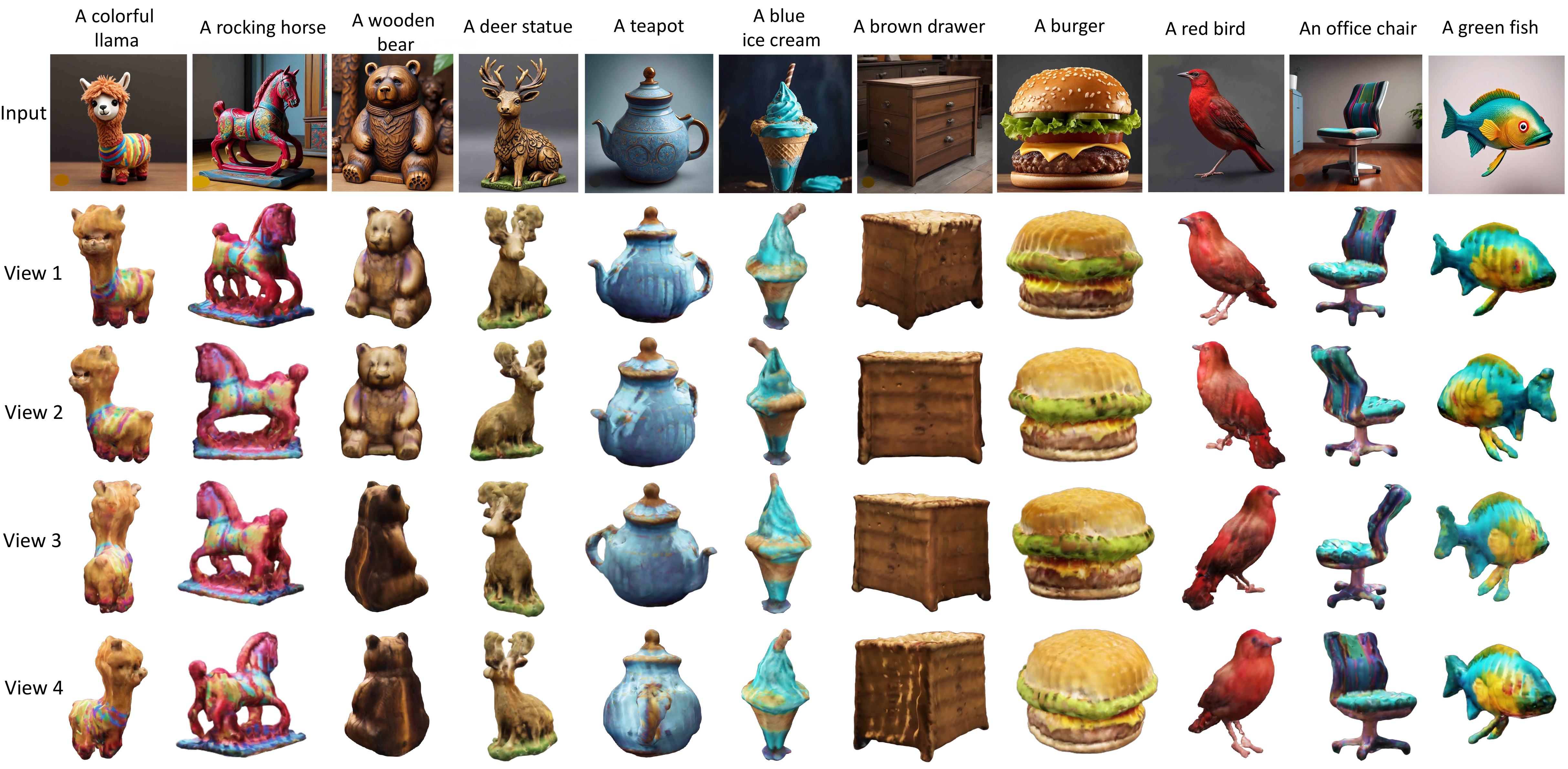}
    \caption{\textbf{Qualitative results of Hyper-Voltran on text-to-3D colored meshes.} The generated images from a diffusion model are used as inputs. We only focus on the main object of the input image.  }
    \label{fig:output_text2img}
\end{figure*}

\subsection{Training and Inference}
Our framework has several losses to train the model, including the HyperNetwork module. Every module is optimized in an end-to-end fashion only in the {training} stage. We define our loss for rendered colors with mean squared error \textit{w.r.t.} the ground-truth $\Vec{C}_i$:
\begin{equation}
    \mathcal{L}_{\textrm{RGB}} = \frac{1}{|P|} \sum^{|P|}_{i=1} \big\| \Vec{\hat{C}}_i-\Vec{{C}}_i \big\|_2^2.
\end{equation}
In addition to the color loss, we also calculate depth predictions supervised with the following loss: 
\begin{equation}
    \mathcal{L}_{\textrm{Depth}} = \frac{1}{|P_1|} \sum^{|P_1|}_{i=1} \big| \Vec{\hat{D}}_i-\Vec{{D}}_i \big|.
\label{eq:depth_loss}
\end{equation}
Also, in order to regularize the SDF
values derived from the SDF network $f_\Psi$, we compute the Eikonal loss~\cite{amos2020implicitgeo} :
\begin{equation}
    \mathcal{L}_{\textrm{Eikonal}} = \frac{1}{|\mathbb{V}|} \sum_{\Vec{v} \in \mathbb{V}} \big(\|\nabla f_\Psi(\Vec{v})\|_2 -1 \big)^2,
    \label{eq:eikonal_loss}
\end{equation}
where $\Vec{v}$ is a sampled 3D point and $\nabla f_\theta(\Vec{v})$ is the gradient relative to the sample point $q$. This loss impacts the surface smoothness.

{Furthermore,}
to empower our framework for generating concise geometric surfaces, we incorporate a sparsity regularization term that penalizes uncontrollable surfaces called a sparse loss~\cite{long2022sparseneus}, expressed as follows: 
\begin{equation}
    \mathcal{L}_{\textrm{Sparse}} = \frac{1}{|\mathbb{V}|} \sum_{\Vec{v} \in \mathbb{V}} \exp \big(-\tau |s(\Vec{v})|\big),
    \label{eq:sparse_loss}
\end{equation}
where $s(\Vec{v})$ is the predicted SDF and $\tau$ is the hyperparameter to scale the SDF prediction. To summarize, The total loss is defined as  $ \mathcal{L}_{\textrm{RGB}} + \mathcal{L}_{\textrm{Depth}} +  \beta_1 \mathcal{L}_{\textrm{Eikonal}} + \beta_2 \mathcal{L}_{\textrm{Sparse}}$. 

\paragraph{Inference.} During inference, there is no more optimization, and only one feed-forward is performed, which reduces the expensive computation to update the models during testing. First, given an input image, we segment the input to extract the foreground object. After we obtain the object with clear background (\eg, white color), we synthesize multi-view scenes from the pretrained Zero123 model~\cite{liu2023zero123} conditioned on the relative change of camera viewpoints. These synthesized images are then employed to generate a 3D mesh by our proposed method. The inference of our proposed method only contains feed-forward, thus comfortably reducing the computational time compared to the existing distillation methods~\cite{poole2022dreamfusion,melaskyriazi2023realfusion,seo20233dfuse}.

\begin{figure*}[t]
    \centering
    \includegraphics[width=.99\textwidth]{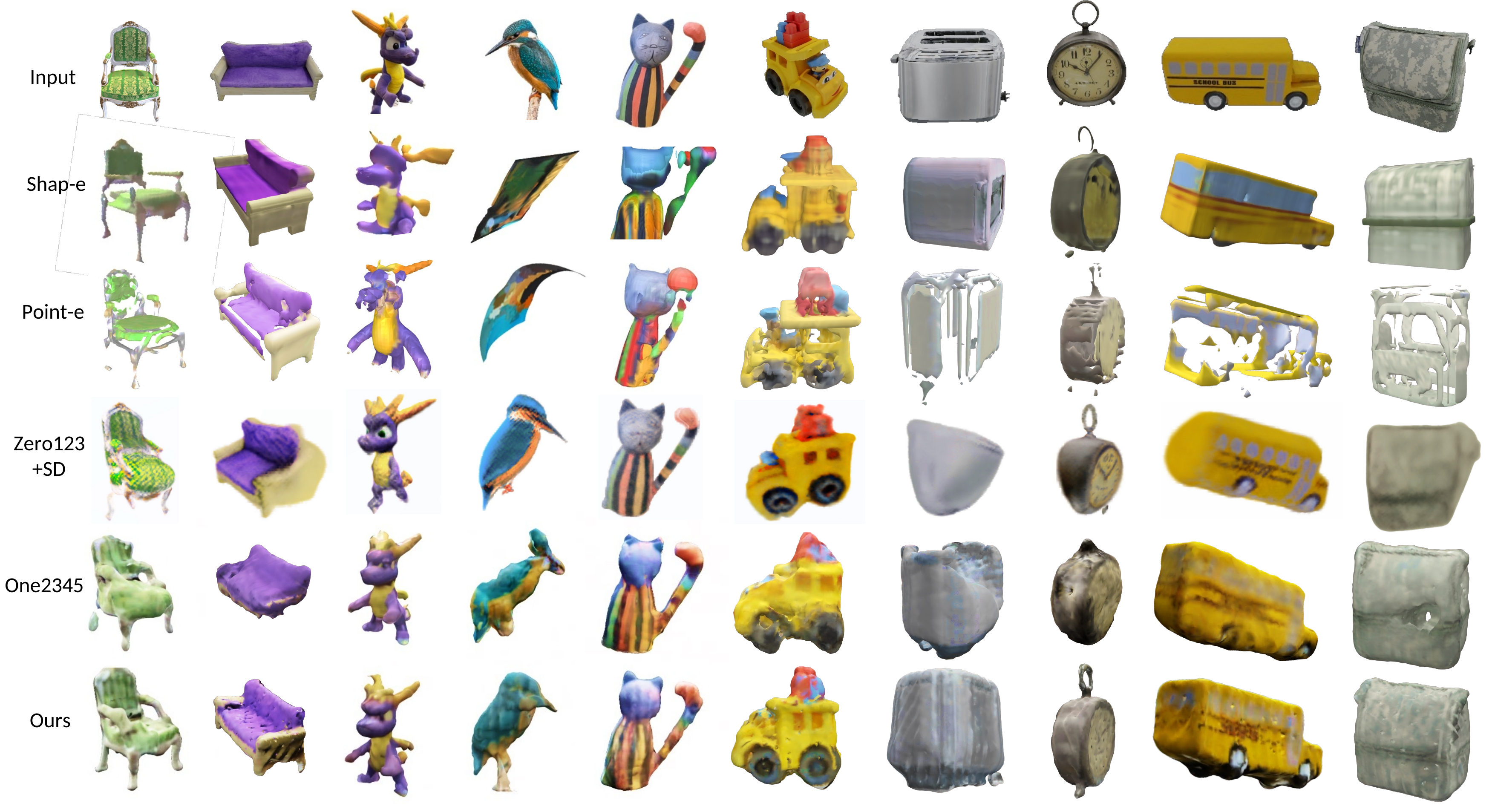}
    \caption{\textbf{Qualitative comparison on single image to 3D reconstruction with previous works} \eg, One2345~\cite{liu2023one2345}, Shap-e~\cite{jun2023shape}, Point-e~\cite{nichol2022pointe}, and Zero123+SD~\cite{poole2022dreamfusion}. VolTran offers more consistent and higher-quality results than competitors, generally providing a higher level of preservation of input details. Please see our supplementary material for more results and zoomed-in details. }
    \label{fig:output_comparison}
\end{figure*}

\section{Experiments}
\label{sec:experiments}

\subsection{Implementation details}
We train our models from publicly available data first shared by~\cite{liu2023one2345}, containing 46K synthesized 3D scenes. 
For the base multi-view generative model, we follow Zero123~\cite{liu2023zero123} and keep its weights frozen.
Additionally, for the geometry-guided encoder, we set the volume encoding size to $96\times 96 \times 96$ for all of our experiments. For the SDF weight generation, we employ the CLIP model~\cite{radford2021clip} as the image encoder, known for generating dependable representations.
In terms of the loss function, we verified that the setting proposed by~\cite{long2022sparseneus} is optimal, \ie, $\beta_1=0.1$ and $\beta_2=0.02$.
On the other hand, during inference, we first apply image segmentation to get an accurate cutout of the target object using the Segment Anything Model (SAM)~\cite{kirillov2023segany}.
Then, we generate 8 key views which are further extended by 4 nearby images each, for a total of 32 viewpoints.

\subsection{Text-to-3D Results}
The text-to-3D pipeline is performed by using off-the-shelf text-to-image models \eg,~\cite{ramesh2022dalle,saharia2022imagen,yu2023cm3leon}. 
We apply the corresponding diffusion process conditioned on a given prompt (\eg, "a wooden bear") and obtain an image depicting it. 
To handle unexpected background information, we cut out the target object from the generated image using SAM~\cite{kirillov2023segany}. 
Different views are further synthesized alongside corresponding camera poses using Zero123~\cite{liu2023zero123}. 
The full set of generated images
are fed to our model, constructing neural encoding volume, generating SDF network weights through a HyperNetwork, and applying global attention, the main components of Hyper-VolTran. 
Fig.~\ref{fig:output_text2img} shows 
results of our method across different views for a given text prompt.
It can be observed from these images that Hyper-Voltran produces good quality meshes that adhere well to corresponding texture, giving a sense of consistency across views.
\begin{figure*}[t]
    \centering
    \includegraphics[width=1.01\textwidth]{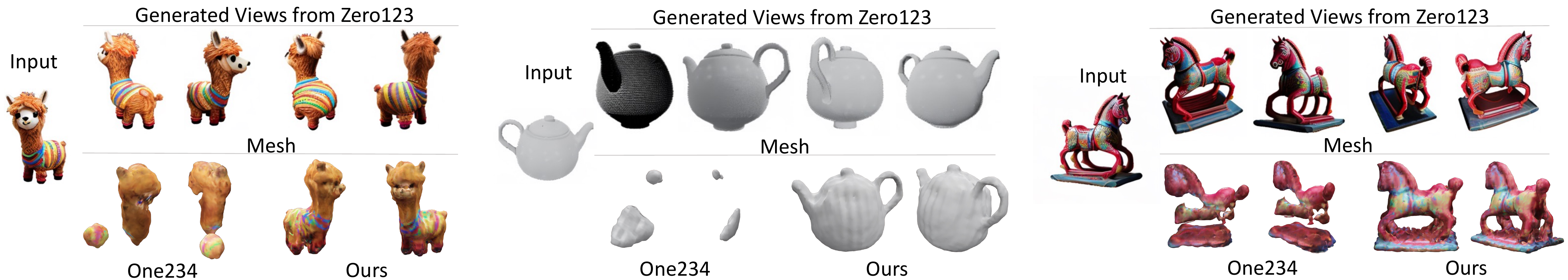}
    \vspace{-5mm}
    \caption{Examples of inconsistently generated views and comparison of our proposed method against One2345~\cite{liu2023one2345} in generating meshes. One2345 fails to build well-reconstructed meshes when the views are arguably inconsistent and challenging.   }
    \label{fig:generated_views}
\end{figure*}

\subsection{Image-to-3D Results}
We use a subset of the GSO dataset~\cite{downs2022gso} to quantitatively evaluate one-shot image-to-3D mesh, comprising 25 objects from different GSO categories. For evaluating rendering quality, we use images from~\cite{melaskyriazi2023realfusion}, spanning 15 objects.

\paragraph{Qualitative results.} We offer qualitative demonstrations of our approach and comparison to One2345~\cite{liu2023one2345}, Shap-e~\cite{jun2023shape}, Point-e~\cite{nichol2022pointe}, and Zero123+SD~\cite{liu2023zero123} in Fig.~\ref{fig:output_comparison}, showcasing Hyper-Voltran's efficacy in addressing one-shot image-to-3D object reconstruction. 
For a fair comparison with One2345~\cite{liu2023one2345}, we employ the same set of synthesized images to generate the 3D meshes. 
We note that One2345~\cite{liu2023one2345} showcases inaccurate and unnatural shapes in Fig.~\ref{fig:output_comparison}. 
Also, we compare to other feed-forward-only approaches~\cite{jun2023shape,nichol2022pointe}. Point-e and Shap-e cannot successfully reconstruct 3D meshes from a single image yielding incorrect colors and shapes. 
Our proposed method is proven robust across a varied set of different objects with higher fidelity and more accurate shapes compared to the baselines.  
We also show in Fig.~\ref{fig:generated_views} some inconsistencies in generated images from Zero123~\cite{liu2023zero123} and how our method can robustly construct the meshes compared to the baseline. 

\paragraph{Quantitative results.} To evaluate our method and compare against baselines in generating meshes, we use the PyTorch3D~\cite{ravi2020pytorch3d} package to calculate Chamfer distance and Iterated Closest Point for source and target alignment to compute F-score. 
In terms of metrics, we follow prior works~\cite{liu2023one2345}, and~\cite{downs2022gso}, 
and use F-Score, Chamfer L2 distance, and intersection-over-union (IoU). 
These metrics are summarized in~Table~\ref{tab:tab_fscore_clipsim}, where Hyper-VolTran proves its improved generalization capabilities on unseen objects by scoring higher than competitors across all tracks, at reasonable computation time cost. 
Similarly, for rendering quality, our method tops all previous works on 3D rendering across all scores: PSNR, LPIPS, and the CLIP similarity score as shown in Table~\ref{tab:tab_clip_psnr}.

\paragraph{Processing Time.} Although our proposed method relies on encoding the input image through an image embedding model and generating weights of the SDF network, the full 3D generation latency is only around 5 seconds on a single A100 GPU. 
This is on par with the processing time of One2345~\cite{liu2023one2345}. 
Additional latency is due to the base diffusion model. 
In our case, we opt to use Zero123~\cite{liu2023zero123} for the synthesis of additional views, 
adding on average around 40 seconds per object. 
As shown in Table~\ref{tab:tab_fscore_clipsim}, the processing time of Shap-e is lower, which results in generally lower quality results than our method. 


\begin{table}[t]
\setlength{\extrarowheight}{4pt}
    \centering
    \resizebox{0.485\textwidth}{!}{
    \Large\addtolength{\tabcolsep}{-2pt}
    \begin{tabular}{c c c c c}
    \hline
         	Method &F-Score ($\uparrow$) &Chamfer L2 ($\downarrow$)	&IoU ($\uparrow$)	
          &Time \\
          \hline
            Point-e~\cite{nichol2022pointe}	& 16.45	&1.73	&0.09 	&78 secs \\
            Shap-e~\cite{jun2023shape}	& 10.10 	&1.98	&0.11 &\textbf{27} secs\\
            Zero123+SD~\cite{liu2023zero123}	&14.85 	 &1.41	&0.21	&15 mins\\
            One2345~\cite{liu2023one2345}	&12.00	&1.90	&0.13	&{45} secs\\
            \rowcolor{blond} \textbf{Hyper-VolTran (ours)}	&\textbf{17.45} &\textbf{1.14} &\textbf{0.22} 		&{45} secs \\
            \hline
    \end{tabular}
    }  
    \vspace{-2mm}
    \caption{F-Score, Chamfer L2, IoU, and time comparison to baselines on the GSO dataset~\cite{downs2022gso}.}
    \label{tab:tab_fscore_clipsim}
\end{table}


\begin{table}[t]
\setlength{\extrarowheight}{2pt}
    \centering
    \resizebox{0.485\textwidth}{!}{
    \Large\addtolength{\tabcolsep}{4.5pt}
    \begin{tabular}{c c c c}
    \hline
         	Method &PSNR ($\uparrow$)	&LPIPS ($\downarrow$) &CLIP Sim. ($\uparrow$) \\	
          \hline
            Point-e~\cite{nichol2022pointe}	&0.98 &0.78	&0.53\\
            Shap-e~\cite{jun2023shape}	&1.23 &0.74	&0.59	\\
            Zero123~\cite{liu2023zero123}	&19.49 &0.11	&0.75	\\
            RealFusion~\cite{melaskyriazi2023realfusion}	&0.67	& 0.14 &0.67\\%
            Magic123~\cite{qian2023magic123}	&19.50	&\textbf{0.10} &0.82\\
            One2345~\cite{liu2023one2345}	&16.10	& 0.32 &0.57	\\
            \rowcolor{blond} \textbf{Hyper-VolTran (ours)}	&\textbf{23.51} &\textbf{0.10} &\textbf{0.86} 		\\
            \hline
    \end{tabular}
    }
    \vspace{-2mm}
    \caption{PSNR, LPIPS, and CLIP similarity comparison to prior works on the collected images in RealFusion~\cite{downs2022gso}.}
    \label{tab:tab_clip_psnr}
\end{table}

\begin{figure}[h]
    \centering
    \includegraphics[width=0.485\textwidth]{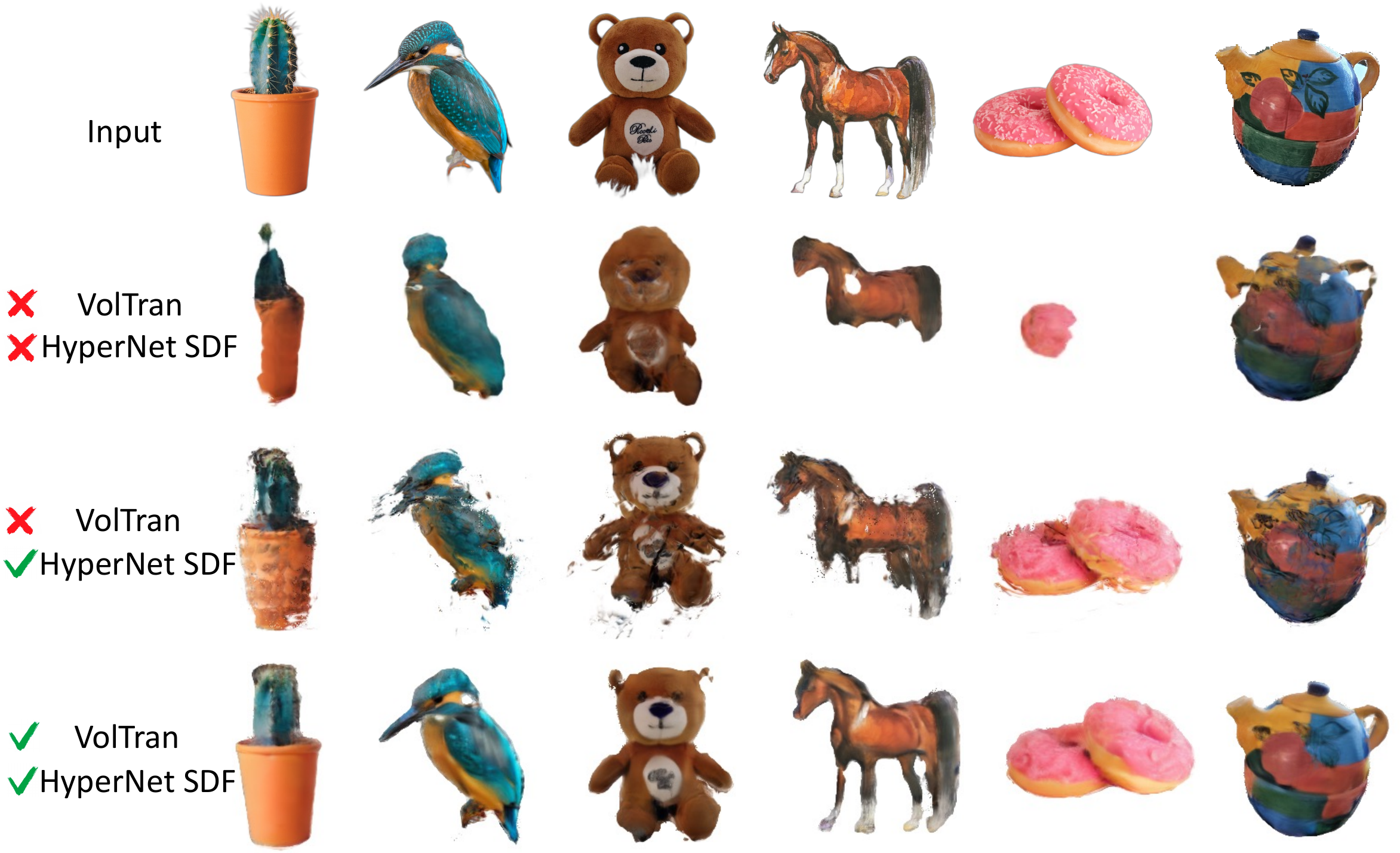}
    \vspace{-5mm}
    \caption{Ablation study on each module. Impacts of each module on rendering colored scenes.  }
    \label{fig:ablation_module}
\end{figure}

\subsection{Analysis and Ablations}
\paragraph{The SDF weight generator via a HyperNetwork and VolTran.}  We investigate the efficacy of our proposed two modules: the HyperNetwork for SDF and VolTran. This ablation study is performed to analyze the impact of each module. As shown in Fig.~\ref{fig:ablation_module}, we can observe that rendering deteriorates without the HyperNetwork and Voltran. While without VolTran, rendering scenes yields some noise as the impact of inconsistent inputs. Using both, we can achieve plausible rendering results.   

\begin{figure}[t]
    \centering
    \includegraphics[width=0.455\textwidth]{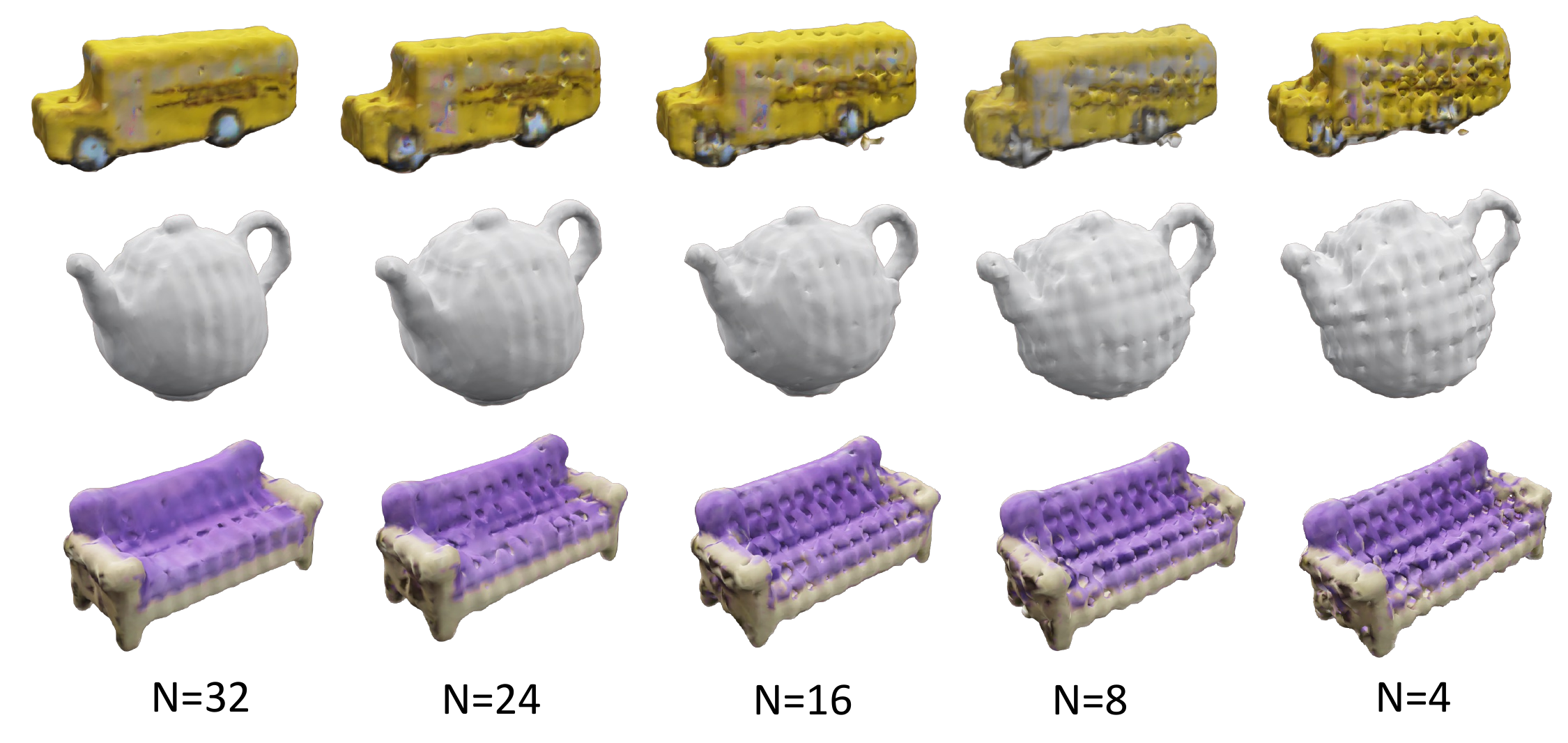}
    \caption{ Qualitative results with different numbers of samples generated from a diffusion model. The more images are generated from the diffusion model, the better shape quality is achieved.   }
    \label{fig:ablation_samples}
\end{figure}

\paragraph{Number of samples.} 
We evaluate the generated results by varying numbers of support images obtained from the diffusion model, ranging from 32 down to 4 images from different perspectives. Fig.~\ref{fig:ablation_samples} showcases the impact of the number of samples generated from the diffusion model. Our approach gains advantages from an increased number of generated images for forming geometry representations. Conversely, an excessively low number of samples leads to degradation.

\section{Conclusions}
\label{sec:Conclusions}
 In this paper, we address the challenge of deriving a 3D object structure from a single image. Our proposed approach, called Hyper-VolTran, comprises a HyperNetwork module and a transformer module. Specifically, HyperNetworks generate SDF weights, while the transformer module facilitates robust global aggregation from inconsistent multi-views. Our method demonstrates effective generalization to unseen objects in the single image-to-3D task, as evidenced by both quantitative and qualitative evaluations. Notably, our approach rapidly generates 3D meshes, accomplishing this task in just 45 seconds without per-scene optimization. Compared with state-of-the-art methods, our proposed approach excels in both time efficiency and reconstruction accuracy.



{
    \small
    \bibliographystyle{ieeenat_fullname}
    \bibliography{main}
}


\end{document}